\documentclass[runningheads]{llncs}

\usepackage{eccv}

\usepackage{eccvabbrv}
\usepackage{graphicx}
\usepackage{placeins}  
\usepackage{booktabs}
\usepackage{amsmath}
\usepackage{amssymb}
\usepackage{bm}
\usepackage{multirow}
\usepackage{url}
\usepackage{xcolor}
\usepackage[accsupp]{axessibility}

\usepackage[pagebackref,breaklinks,colorlinks,citecolor=eccvblue]{hyperref}
\usepackage{orcidlink}
\newcommand{\projectleader}{\textsuperscript{\ensuremath{\dagger}}}

\begin{document}

\title{Drift-AR: Single-Step Visual Autoregressive Generation via Anti-Symmetric Drifting}
\titlerunning{Drift-AR: Single-Step Visual AR via Drifting}

\author{Zhen Zou\inst{1} \and Xiaoxiao Ma\inst{1} \and Mingde Yao\inst{2} \and Jie Huang\inst{3}\projectleader \and LinJiang Huang \inst{4} \and Feng Zhao\inst{1}}
\authorrunning{Z.~Zou et al.}
\institute{University of Science and Technology of China \and The Chinese University of Hong Kong, MMLAB \and
JD Explore Academy \and Beihang University \\
\email{zouzhen@mail.ustc.edu.cn}}

\maketitle
\begingroup
\renewcommand{\thefootnote}{\fnsymbol{footnote}}
\footnotetext[4]{Project leader.}
\endgroup

\begin{abstract}
Autoregressive (AR)-Diffusion hybrid paradigms combine AR's structured semantic modeling with diffusion's high-fidelity synthesis, yet suffer from a dual speed bottleneck: the sequential AR stage and the iterative multi-step denoising of the diffusion vision decode stage. Existing methods address each in isolation without a unified principle design.
We observe that the per-position \emph{prediction entropy} of continuous-space AR models naturally encodes spatially varying generation uncertainty, which simultaneously governing draft prediction quality in the AR stage and reflecting the corrective effort required by vision decoding stage, which is not fully explored before.
Since entropy is inherently tied to both bottlenecks, it serves as a natural unifying signal for joint acceleration. 
In this work, we propose \textbf{Drift-AR}, which leverages entropy signal to accelerate both stages:
1) for AR acceleration, we introduce Entropy-Informed Speculative Decoding that align draft--target entropy distributions via a causal-normalized entropy loss, resolving the entropy mismatch that causes excessive draft rejection;
2) for visual decoder acceleration, we reinterpret entropy as the \emph{physical variance} of the initial state for an anti-symmetric drifting field---high-entropy positions activate stronger drift toward the data manifold while low-entropy positions yield vanishing drift---enabling single-step (1-NFE) decoding without iterative denoising or distillation.
Moreover, both stages share the same entropy signal, which is computed once with no extra cost.
Experiments on MAR, TransDiff, and NextStep-1 demonstrate 3.8--5.5$\times$ speedup with genuine 1-NFE decoding, matching or surpassing original quality. Code will be available at https://github.com/aSleepyTree/Drift-AR.
\end{abstract}

\section{Introduction}
\label{sec:intro}

\begin{figure}[!tbp]
\centering
\includegraphics[width=.9 \linewidth]{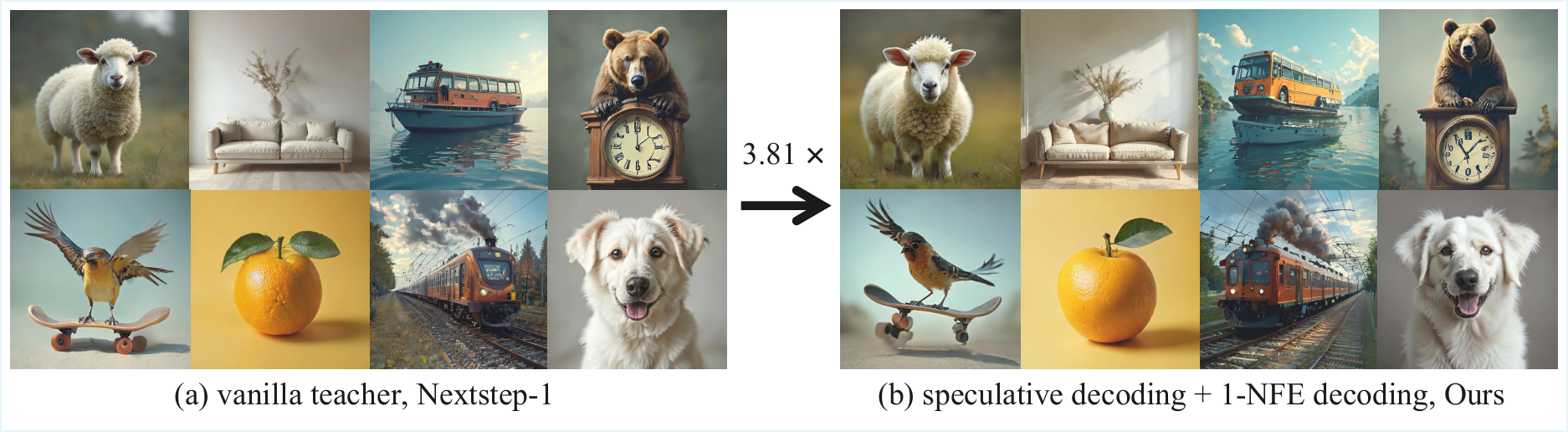}
\caption{Qualitative generation comparison between the vanilla NextStep-1~\cite{team2025nextstep} and our method on GenEval~\cite{ghosh2023geneval}.}
\vspace{-0.5cm}
\label{fig1:show}
\end{figure}

Autoregressive (AR) models have achieved remarkable success in large language models (LLMs)~\cite{achiam2023gpt4,anil2023palm,touvron2023llama}, inspiring extensions to visual generation~\cite{liu2024lumina_mgpt,sun2024llamagen}. To overcome the quality ceiling imposed by discrete tokenization~\cite{van2017vqvae,esser2021vqgan}, recent hybrid AR-Diffusion paradigms~\cite{li2024mar,zhen2025marrying,ren2024flowar,team2025nextstep} shift generation to continuous latent spaces with two stages, combining AR stage's structured semantic modeling with diffusion decoder stage's high-fidelity synthesis~\cite{sohl2015deep,ho2020denoising,song2020score}. These hybrids have achieved impressive results, balancing semantic coherence and visual realism without the information loss inherent to vector quantization.

However, the two-stage design introduces a \emph{dual speed bottleneck}.
In this hybrid inference paradigm, the AR transformer generates tokens sequentially. Each generated AR latent is then decoded by the diffusion decoder through iterative multi-step denoising.
Existing acceleration methods treat these two bottlenecks independently: 1) speculative decoding~\cite{li2024eagle,li2024eagle2} for the AR acceleration and, 2) typical distillation~\cite{yin2024one,yin2024improved} for the visual decoder acceleration. This isolated treatment lacks a unified guiding principle and leaves substantial room for their joint acceleration. 
Moreover, distillation methods such as consistency or distribution-matching distillation (CD/DMD)~\cite{song2023consistency,yin2024one,yin2024improved} remain tied to the multi-step diffusion paradigm: although they reduce the number of denoising steps, iterative inference is still required.

\begin{figure}[!tbp]
\centering
\includegraphics[width=.75\linewidth]{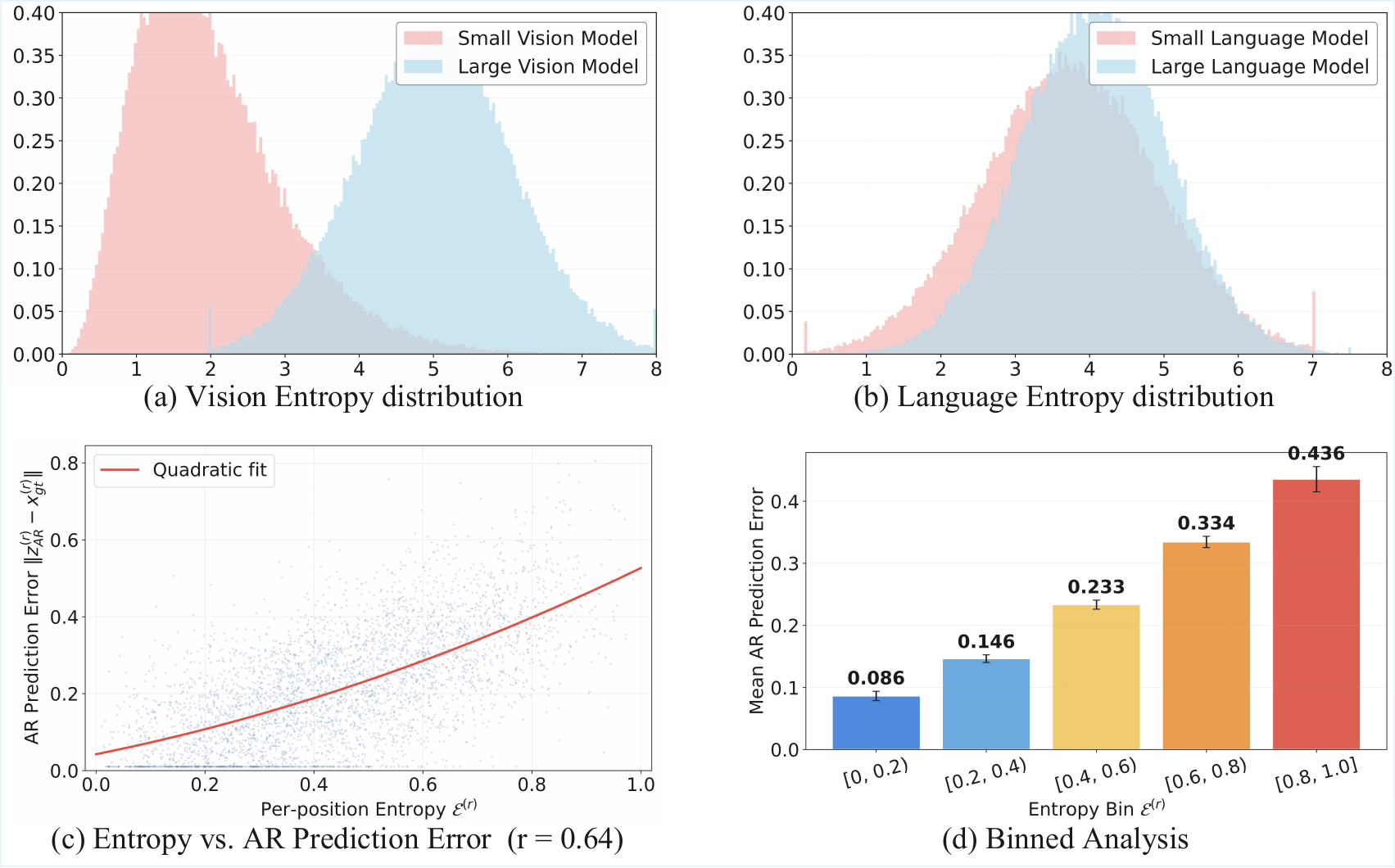}
\caption{Entropy as a diagnostic signal for AR-Diffusion hybrids. (a)~Vision AR entropy: the draft model (red) concentrates at low entropy while the target model (blue) spans higher values, revealing severe \emph{entropy mismatch}. (b)~Language AR entropy: large and small models overlap substantially, explaining why speculative decoding succeeds in LLMs, which cannot be directly applied for vision AR models. (c)~Per-position AR prediction error $\|z_{AR}^{(r)}{-}x_{gt}^{(r)}\|$ vs.\ entropy $\mathcal{E}^{(r)}$ (Pearson $r{=}0.64$): higher entropy correlates with larger prediction error, confirming that entropy encodes local generation difficulty. (d)~Binned analysis: mean AR error increases monotonically with entropy, motivating entropy-parameterized variance for the drifting decoder.}
\label{fig2:moti}
\end{figure}

We start our work to accelerate the AR-Diffusion hybrids generation by making a key observation: unlike LLMs generation, real-world images exhibit highly non-uniform information distribution, leading to heterogeneous generation uncertainty across spatial positions.
In continuous-space AR models, the \emph{per-position predicted entropy} naturally captures this heterogeneity, providing a signal with deep implications for both AR acceleration and diffusion visual decoding acceleration, which is not fully explored by existing methods.
Specifically, redundant regions (\eg, sky, blank walls) tend to induce low entropy, whereas complex structures (textures, object boundaries) produce high entropy. 
Crucially, this entropy carries a \emph{dual meaning} for the dual speed bottleneck: (1)  it reflects the reliability of draft predictions in speculative decoding: entropy mismatch between small draft models and target large models often leads to excessive rejection (see Fig.~\ref{fig2:moti}(a)); and (2)~it encodes the local generation difficulty at each spatial position, indicating how much ``corrective effort'' the diffusion-derived visual decoder should apply. We therefore argue that entropy forms a \emph{natural bridge} that connects AR acceleration and diffusion visual decoding acceleration within a unified framework.

Based on the above observation and analysis, we first address the AR acceleration bottleneck via \textbf{Entropy-Informed Speculative Decoding} (see left in Fig.~\ref{fig3:method}). Directly applying speculative decoding approach (e.g., EAGLE~\cite{li2024eagle}) to continuous-space AR yields poor acceptance rates, as smaller draft models tend to produce low-entropy, overconfident features that deviate from the large target's entropy distribution (see Fig.~\ref{fig2:moti}(a)). This mismatch arises from the low information density of visual data, where the limited capacity of the draft model causes predictions to collapse toward dominant modes. To mitigate this issue, we adapt EAGLE to continuous latent spaces with two key modifications: (i)~a continuous-feature regression loss that replaces the discrete classification objective that matches the vision AR model's characteristics, and 
(ii)~a causal-normalized entropy loss that explicitly aligns the entropy distributions of the draft and target models, thereby improving acceptance rates of speculative decoding in AR acceleration.

We then address the visual decoding bottleneck by going beyond entropy as a simple confidence signal (see middle and right in Fig.~\ref{fig3:method}). We replace the iterative diffusion-derived decoder with an \textbf{anti-symmetric drifting field}~\cite{deng2026drifting} and reinterpret the AR prediction entropy as the \emph{physical variance} of the drifting process's initial distribution. This leads to an \textbf{Entropy-Parameterized Prior}: at each spatial position, the AR feature serves as the prior mean, while the entropy-derived variance determines the spread of the initial state.
Empirically, this formulation is well supported. As shown in Fig.~\ref{fig2:moti}(c)(d), per-position AR prediction error strongly correlates with entropy (Pearson $r{=}0.64$), indicating that high-entropy regions correspond to locations where AR predictions deviate most from the ground truth and therefore require stronger correction during decoding.
The formulation is physically consistent with the anti-symmetric drifting field ($V_{p,q}{=}{-}V_{q,p}$): high-entropy positions induce larger variance and thus stronger drift toward the data manifold, while low-entropy positions yield near-zero variance where the drift naturally vanishes ($q{\approx}p \Rightarrow V{\to}0$). This enables principled single-step generation for visual decoding stage without iterative refinement.

The two components are naturally unified through the shared entropy signal: the per-position entropy $\mathcal{E}_{\text{entropy}}^{(r)}$ computed once during speculative decoding is directly reused as the variance parameter for the visual decoder's initial distribution, requiring no additional computation or separate uncertainty estimation. 
Therefore, the proposed framework is end-to-end training, which is enabled by an annealed schedule that first stabilizes the AR entropy, then optimizes the drifting field over a fixed prior. Unlike CD/DMD-based acceleration, our framework eliminates the need for a multi-step teacher and avoids distillation instability, achieving 1-NFE visual decoding from first principles.

Our contributions are:
\begin{itemize}
\item We identify the \emph{dual role} of prediction entropy in continuous-space AR-Diffusion hybrids, which governs both draft quality in speculative decoding for AR stage and local generation difficulty for visual decoding stage, and thus identify entropy signal to unify the two stage's joint acceleration.
\item We introduce Drift-AR, which couples entropy-informed speculative decoding for AR stage acceleration, along with an entropy-parameterized anti-symmetric drifting field for visual decoding stage acceleration, achieving genuine single-step (1-NFE) visual generation without iterative denoising or distillation.
\item Extensive experiments on MAR, TransDiff, and NextStep-1 demonstrate 3.8--5.5$\times$ speedup while matching or surpassing original generation quality and diversity.
\end{itemize}

\FloatBarrier
\section{Related Work}
\label{sec:related}

\subsection{AR-Diffusion Hybrid Paradigm for Generation}

The AR-Diffusion hybrid bridges AR's structured semantic modeling and diffusion's high-fidelity synthesis. MAR~\cite{li2024mar} introduced a diffusion-inspired denoising loss on continuous latent features, avoiding discrete vector quantization while preserving AR's sequential prediction. TransDiff~\cite{zhen2025marrying} achieved end-to-end integration by coupling an AR transformer with a DiT-based diffusion decoder, mitigating AR's $O(n^2)$ complexity. FlowAR~\cite{ren2024flowar} replaced diffusion with flow matching~\cite{lipman2022flow,liu2022flow} and added Spatial-adaLN for feature alignment, while NextStep-1~\cite{team2025nextstep} unified AR and flow matching in a single transformer. These models share core principles: AR supplies hierarchical semantic guidance, diffusion/flow matching handles high-dimensional visual synthesis, and tight integration via joint training or adaptive feature injection preserves guidance--synthesis alignment.

\subsection{Acceleration of Generative Models}

Acceleration techniques for generative models target the distinct bottlenecks of AR (sequential latency) and diffusion (iterative denoising)~\cite{tian2024var,wang2024par,bolya2023token,bolya2022token,lou2024token,yuan2024ditfastattn,Structuralpruning,zhang2024laptop,shang2023post,so2024temporal}. For diffusion, distillation-based methods such as DMD~\cite{yin2024one} and DMD2~\cite{yin2024improved} align student--teacher distributions to reduce denoising to one or few steps, while SDXL-Lightning~\cite{lin2024sdxl} combines progressive distillation~\cite{salimans2022progressive} with adversarial loss. Other works train few-step or single-step generators directly~\cite{deng2026drifting}. However, synergistic acceleration for AR-Diffusion hybrids remains largely unaddressed, with current methods optimizing each component in isolation.

\FloatBarrier
\section{Preliminary}
\label{sec:preliminary}

\subsection{AR-Diffusion Hybrid Paradigm}
As an AR-Diffusion hybrid model, MAR~\cite{li2024mar} operates directly on continuous latents $x \in \mathbb{R}^{d}$ with a diffusion-inspired objective:
\begin{equation}
\mathcal{L}_{MAR} = \mathbb{E}_{\epsilon \sim \mathcal{N}(0,I), t \sim \text{Uniform}(0,1)}\left[\|\epsilon - \epsilon_\theta(x_t \mid t, z_{AR})\|^2\right],
\end{equation}
where $x_t = \sqrt{\bar{\alpha}_t}x + \sqrt{1-\bar{\alpha}_t}\epsilon$ is the noisy latent, $z_{AR}$ denotes AR's predicted feature, and $\epsilon_\theta$ represents the diffusion noise estimator.

TransDiff~\cite{zhen2025marrying} further enables end-to-end integration: its AR transformer models sequential semantic features as $z = \text{ART}(x_{1:i-1}, c)$, where $c$ is class/text condition and $x_{1:i-1}$ are preceding latents, \text{ART} is the AR diffusion transformer. A DiT-based diffusion decoder $\text{Diff}_\theta$ then reconstructs images via: $x_0 = \text{Diff}_\theta(x_t \mid t, z)$. Although TransDiff is capable of generating all tokens in one step, we treat it as a multi-step method for improved performance.

\subsection{Acceleration}
Speculative decoding accelerates AR generation via a ``draft--verify'' mechanism: a lightweight draft model proposes predictions in parallel, which a stronger target model verifies. EAGLE~\cite{li2024eagle} and its successors~\cite{li2024eagle2,li2025eagle} are the state-of-the-art for \emph{discrete-token} NLP models, employing feature regression and classification objectives:
\begin{equation}
\label{eq_loss_reg_cls}
\begin{aligned}
\mathcal{L}_{\text{reg}} &= \text{Smooth L1}(f_{j+1}, \text{Draft Model}(T_{2:j+1}, F_{1:j})), \\
\mathcal{L}_{\text{cls}} &= \text{Cross-Entropy}(p_{j+2}, \hat{p}_{j+2}),
\end{aligned}
\end{equation}
where $F_{1:j}$ denotes draft model's current feature sequence, $\hat{f}_{j+1}$ the target model's next-step feature, $T_{2:j+1}$ the output tokens, and $p/\hat{p}$ the corresponding discrete probability distributions. Adapting EAGLE to the continuous-space AR-Diffusion setting requires non-trivial modifications, as we discuss in Sec.~\ref{sec:method}.

In contrast, \emph{drift-based} generative models~\cite{deng2026drifting} avoid time-step-based iterative denoising. A network $f_\theta$ maps noise $\epsilon \sim p_\epsilon$ to samples $x = f_\theta(\epsilon)$; the pushforward distribution $q = (f_\theta)_{\#} p_\epsilon$ is evolved during training via a \textbf{drifting field} $V_{p,q}(x)$ so that $q$ approaches the data distribution $p$. When $V$ is \textbf{anti-symmetric} ($V_{p,q}=-V_{q,p}$), equilibrium $q=p$ implies $V=0$, enabling \textbf{single-step (1-NFE)} generation at inference without iterative refinement. This avoids distillation-based acceleration (e.g., DMD~\cite{yin2024one}) and its need for a multi-step teacher.

\FloatBarrier
\section{Method}
\label{sec:method}

\begin{figure}[!tbp]
\centering
\includegraphics[width=\linewidth]{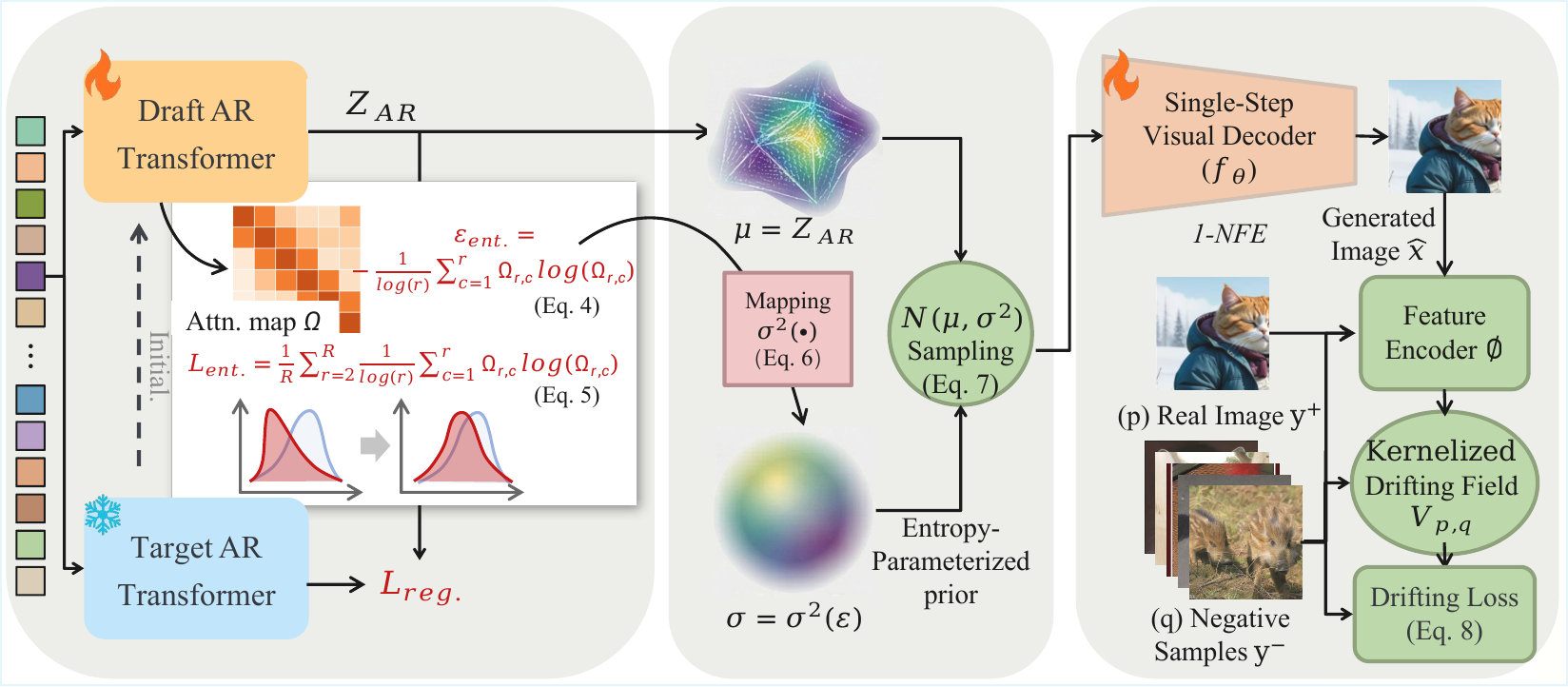}
\caption{Illustration of the proposed Drift-AR framework. (Left) Entropy-informed speculative decoding alleviates entropy mismatch between draft AR model and target AR model, which provides entropy-aligned semantic guidance that drives the draft AR model learns diverse, uncertainty-aware feature predictions rather than collapsing to overconfident modes. (Right) The visual decoder learns an anti-symmetric drifting field $V_\theta$ guided by Entropy-Parameterized Prior over the pushforward distribution $q$; the training procedure evolves $q$ toward the data distribution $p$, thus when equilibrium the drift vanishes and \emph{single-step} (1-NFE) generation is achieved. The overall Drift-AR framework is end-to-end training, which couples entropy-guided AR with drifting-field vision decoder optimization.}
\label{fig3:method}
\end{figure}

As shown in Fig.~\ref{fig3:method}, Drift-AR is organized around a single guiding signal---the per-position prediction entropy $\mathcal{E}_{\text{entropy}}^{(r)}$---which drives both AR acceleration and single-step visual decoding. First, entropy informs speculative decoding by aligning draft--target entropy distributions (Sec.~\ref{sec:spec}). The same entropy is then reinterpreted as the physical variance of an entropy-parameterized prior for an anti-symmetric drifting field, enabling 1-NFE generation (Sec.~\ref{sec:drifting}). A unified pipeline ties both components together (Sec.~\ref{sec:pipeline}).

\subsection{Entropy-Informed Speculative Decoding}
\label{sec:spec}
Speculative decoding is a natural candidate for accelerating the AR stage of AR-Diffusion models, but direct application of EAGLE~\cite{li2024eagle} yields poor performance. The draft model's outputs are frequently rejected by the target, and its low-entropy (overconfident) predictions lead to diminished diversity (Fig.~\ref{fig2:moti}(a)). The root cause is \emph{entropy mismatch}: the small model produces overly confident features that deviate from the target's entropy distribution. As shown in the left of Fig.~\ref{fig3:method}, we adapt EAGLE to the continuous latent space with two changes.

\noindent\textbf{Continuous-space regression loss.} We remove EAGLE's discrete classification loss $\mathcal{L}_{\text{cls}}$ (Eq.~\ref{eq_loss_reg_cls}) and replace its token-conditioned regression with a continuous-feature variant. The draft is conditioned on the continuous AR feature sequence rather than discrete output tokens:
\begin{equation}
\mathcal{L}_{\text{reg}} = \text{Smooth\,L1}\!\left(z_{j+1},\; \text{DraftModel}(Z_{1:j},\, F_{1:j})\right),
\label{eq:reg_continuous}
\end{equation}
where $Z_{1:j}$ is the sequence of continuous AR features produced so far, $F_{1:j}$ is the target model's hidden-state sequence, and $z_{j+1}$ is the target's next-step feature. This preserves structural alignment between draft and target while remaining consistent with continuous-space AR.

\noindent\textbf{Causal-normalized entropy loss.} In a causal-masked attention matrix, row $r$ attends only to row $1,\ldots,r$, so its maximum possible entropy is $\log(r)$. A plain average over rows would unfairly penalize early rows (e.g., $r{=}1$ has entropy zero by construction) and distort gradients. We normalize each row's negative entropy by $\log(r)$:
\begin{equation}
\mathcal{E}_{\text{entropy}}^{(r)} \triangleq -\frac{1}{\log(r)}\sum_{c=1}^{r}\Omega_{r,c}\log\Omega_{r,c},
\label{eq_entropycomp}
\end{equation}
where per-position entropy indicator $\mathcal{E}_{\text{entropy}}^{(r)} \in [0,1]$, and total entropy loss is calculated as:
\begin{equation}
\mathcal{L}_{\text{entropy}} = -\frac{1}{R}\sum_{r=2}^{R}\mathcal{E}_{\text{entropy}}^{(r)},
\label{eq_entropyloss}
\end{equation}
where $\Omega_{r,c}$ is the $(r,c)$-th softmax entry of the penultimate attention map ($\Omega_{r,c}{=}0$ for $c{>}r$; we start at $r{=}2$ since row $r{=}1$ has a single support and zero entropy). Minimizing $\mathcal{L}_{\text{entropy}}$ maximizes the \emph{normalized} attention entropy at every position, so that training and the length-normalized inference threshold (Sec.~\ref{sec:pipeline}) use the same scale.

Note that $\mathcal{E}_{\text{entropy}}^{(r)}$ has a dual role: it regularizes the draft to match the target's entropy during speculative decoding, and it is reused as the \emph{physical variance} of the $r$-th AR feature when building the initial distribution for the drifting decoder (Sec.~\ref{sec:drifting}), tying AR uncertainty directly to the generative process.

\subsection{Entropy-Parameterized Single-Step Generative Drifting}
\label{sec:drifting}
In typical AR-Diffusion hybrids (e.g., MAR, TransDiff, NextStep-1), the visual decoder is conditioned on AR features but its initial state is drawn from a fixed Gaussian prior unrelated to the AR prediction's confidence; the visual decoder does not treat AR uncertainty as part of the generative state.

We instead treat the AR output as the \emph{physical} initial distribution for a single-step drifting process: the prior $q$ is defined directly from the AR features, with variance tied to per-position entropy, where entropy is related to the determinacy of generation---higher entropy indicates greater uncertainty like what we have shown in Fig.~\ref{fig2:moti}(c)(d), so that confident predictions yield low-variance initial states and uncertain ones yield higher variance.

\noindent\textbf{Mapping $\mathcal{E}_{\text{entropy}}^{(r)}$ to vision decoder initial distribution.}
As shown in the middle of Fig.~\ref{fig3:method}, we first map the normalized per-position entropy 
$\mathcal{E}_{\text{entropy}}^{(r)} \in [0,1]$ 
to a non-negative variance scale.
Since entropy reflects structural uncertainty in the AR process,
we use a monotonic mapping so that higher entropy corresponds to larger variance.
Specifically, we adopt a bounded exponential temperature mapping:
\begin{equation}
    \sigma(\mathcal{E}) 
    = \sigma_{\max} 
    \cdot 
    \frac{e^{\mathcal{E}/\tau_\sigma} - 1}
         {e^{\mathcal{E}_{\max}/\tau_\sigma} - 1},
    \label{eq:sigma_mapping}
\end{equation}
where $\sigma_{\max}$ caps the variance,
$\tau_\sigma$ controls curvature,
and $\mathcal{E}_{\max}$ normalizes the input range to $[0,\sigma_{\max}]$
($\sigma_{\max}{=}0.5$, $\tau_\sigma{=}2.0$ in experiments).
When $\mathcal{E}$ is small (confident AR),
$\sigma(\mathcal{E})$ remains close to zero;
as $\mathcal{E}$ approaches 1,
the variance smoothly increases toward $\sigma_{\max}$.

The $R{=}h{\times}w$ AR features are reshaped into a 2D entropy map $\mathbf{E} \in \mathbb{R}^{h \times w}$ with $\mathbf{E}_{i,j} = \mathcal{E}_{\text{entropy}}^{(r)}$ ($r{=}i{\cdot}w{+}j$). At each position $r$, we construct a Gaussian prior centered at the AR feature with entropy-determined variance:
\begin{equation}
    q\!\left(x^{(r)} \mid z_{AR}^{(r)}, \mathcal{E}_{\text{entropy}}^{(r)}\right)
    =
    \mathcal{N}\!\left(
        x^{(r)};
        z_{AR}^{(r)},
    \sigma^2\!\left(\mathcal{E}_{\text{entropy}}^{(r)}\right)\mathbf{I}
    \right).
\end{equation}
Thus, positions with low entropy (confident AR grounding)
produce sharply concentrated initial states,
whereas high-entropy positions yield broader initial distributions.

We sample $x_0^{(r)}$ from $q^{(r)}$ via the standard Gaussian reparameterization
$x_0^{(r)} = z_{AR}^{(r)} + \sigma(\mathcal{E}_{\text{entropy}}^{(r)})\,\epsilon$,
where $\epsilon \sim \mathcal{N}(0,I)$,
yielding the initial decoder state
$x_0 \in \mathbb{R}^{h \times w \times d}$.

\noindent\textbf{Aligning AR latents with image feature via drifting field.} 
As shown in the right of Fig.~\ref{fig3:method}, we train a visual decoder $f_{\theta}$ with $x_0$ sampled from the prior $q$ as input and $c$ as the condition. The entropy map $\mathbf{E}$ is injected via additional Entropy-AdaLN. The corresponding output is
$\hat{\mathbf{x}} = f_{\theta}(x_0, c, \mathbf{E})$, through which the decoder aligns the AR latents with the VAE-constrained image latent space.

To avoid dimension mismatch between the decoder's latent output and the drifting field defined in feature space,
we map $\hat{\mathbf{x}}$ into the corresponding latent space using a frozen feature encoder $\phi$. For simplicity and effectiveness, we use a copy of the target AR model's penultimate layer as $\phi$, which is fixed throughout training.
The loss is therefore formulated entirely in $\phi$-space, following the kernelized attraction--repulsion construction of the drifting field $V_{p,q}$:
\begin{equation}
    \mathcal{L}_{\text{drift}} = \mathbb{E}_{x \sim q}\left[ \left\| \phi(\hat{\mathbf{x}}) - \text{stopgrad}\Big( \phi(\hat{\mathbf{x}}) + V_{p,q}\big(\phi(\hat{\mathbf{x}})\big) \Big) \right\|_2^2 \right].
    \label{eq:drift_loss}
\end{equation}

Specifically, with prediction and regression target constrained in same feature space, $V_{p,q}$ is computed following drifting model~\cite{deng2026drifting}: for each generated sample, real samples exert an \emph{attraction} with field $V_p^+$, and generated samples \emph{repulsion} with $V_q^-$, where $p,q$ represents real data and generated samples, respectively.
With kernel functions at multiple temperatures for characterizing similarities among positive and negative samples, the net force yields the drift direction.

Gradients flow through $\phi(\hat{\mathbf{x}})$ and the decoder $f_{\theta}$,
while $\phi$ itself remains frozen.
This design is physically self-consistent:
high-entropy positions receive larger variance and therefore stronger drift,
whereas low-entropy positions have near-zero variance where the field naturally vanishes.
This property enables 1-NFE generation without iterative refinement.

\subsection{Unified Training and Inference Pipeline}
\label{sec:pipeline}

\noindent\textbf{Annealed training for stabilizing the drifting field.}
A naive end-to-end training that jointly updates the AR module and the drifting visual decoder suffers from the ``moving prior'' problem: since the source distribution $q$ is induced by the AR outputs, updating the AR module changes $q$ at every iteration. As a result, the drifting field is effectively asked to transport samples toward a continuously shifting target, which makes the drifting dynamics unstable.

This is particularly problematic because the anti-symmetric drifting field $V_{p,q}$ is well-defined only when both the source $q$ and the target $p$ are stationary, in which case $q{=}p \Rightarrow V_{p,q}{=}0$ corresponds to a true equilibrium. When $q$ keeps moving, this equilibrium interpretation breaks.
We therefore adopt a two-phase schedule controlled by a scalar weight $\alpha(t)$:
\begin{equation}
\alpha(t) = \begin{cases}
\alpha_0 - \alpha_0 \cdot \dfrac{t}{T_{\text{freeze}}}, & t \leq T_{\text{freeze}},\\[6pt]
0, & t > T_{\text{freeze}},
\end{cases}
\end{equation}
where $\alpha_0=0.95$ and $T_{\text{freeze}}= 0.8 \cdot T_{\text{total}}$. The total loss is:
\begin{equation}
\mathcal{L}_{\text{total}} = \alpha(t) \cdot (\mathcal{L}_{reg} + \mathcal{L}_{entropy}) + (1 - \alpha(t)) \cdot \mathcal{L}_{\text{drift}},
\end{equation}
with $\mathcal{L}_{\text{drift}}$ (Eq.~\ref{eq:drift_loss}), 
$\mathcal{L}_{\text{reg}}$ (Eq.~\ref{eq:reg_continuous}), 
and $\mathcal{L}_{\text{entropy}}$ (Eq.~\ref{eq_entropyloss}). 
Accordingly, the training process is divided into two phases:

\noindent\textit{Phase I ($t \leq T_{\text{freeze}}$): Joint optimization.}
$\alpha(t)$ decays linearly from $\alpha_0$ to $0$.
We initialize $\alpha_0{<}1$ ($0.95$ rather than $1$) so that the drifting decoder receives a small supervisory signal from the beginning, avoiding cold-start collapse.
During this phase, the AR prior is optimized with $\mathcal{L}_{\text{reg}}$ and $\mathcal{L}_{\text{entropy}}$, while the drifting decoder simultaneously learns to correct the evolving source distribution $q$ toward the target distribution $p$.
By the end of Phase~I, the AR-induced prior $q$ becomes sufficiently stable for subsequent drifting-field optimization.

\noindent\textit{Phase II ($t > T_{\text{freeze}}$): Pure drifting.} $\alpha(t)=0$, so only $\mathcal{L}_{\text{drift}}$ remains active.
To keep the source distribution $q$ stationary, the AR prior is frozen in Phase~II when constructing $x_0$, preventing $\mathcal{L}_{\text{drift}}$ from back-propagating into the AR module and avoiding potential gradient leakage introduced by the reparameterization (Sec.~\ref{sec:drifting}).
The drifting decoder then optimizes $V_{p,q}$ over a fixed source distribution $q$, restoring the stationary-source assumption required by the equilibrium condition of drifting dynamics.

\noindent\textbf{Context-aware early-stopping for efficient speculation.} During speculative decoding we monitor the draft's attention entropy $\mathcal{E}_{\text{entropy}}^{(r)}$ to decide when to stop speculation and fall back to the target.
We use the \emph{shallow}-layer entropy~\cite{wang2024continuous}, which correlates with prediction quality and is computationally efficient.
When the normalized entropy falls below a threshold, speculation stops and the target model verifies the draft prefix in parallel~\cite{leviathan2023fast}. The target accepts the longest matching prefix, discards the rest, and continues autoregressive generation from the accepted position.
This avoids wasting target forward passes on low-quality draft segments.
The threshold $\tau_{\text{global}}$ must handle two issues:

\noindent\emph{Length normalization.}
Based on the normalized entropy $\mathcal{E}_{\text{entropy}}^{(r)}$ at position $r$
(denoted as $\mathcal{E}^{r}$) in Eq.~\ref{eq_entropycomp},
we set a global baseline
$\tau_{\text{global}} = 0.3\,\mathbb{E}[\mathcal{E}^{r}] - 0.1\,\text{std}(\mathcal{E}^{r})$
from training statistics,
so that the draft acceptance policy is neither overly conservative nor aggressive,
achieving a better balance between generation efficiency and quality.

\noindent\emph{Content-aware adaptation.}
A constant $\tau_{\text{global}}$ can trigger premature or late stopping in inherently low-entropy regions (e.g., smooth sky).
To make the stopping criterion content-aware, we set the threshold based on the target model's normalized entropy.
Specifically, at context length $c$ we compute the target entropy $\hat{\mathcal{E}}^{\text{target}}_{c}$ and maintain an EMA $\bar{\mathcal{E}}^{r}_\text{target} = 0.9\,\bar{\mathcal{E}}^{r}_\text{target} + 0.1\,\mathcal{E}^{r}_\text{target}$ with initialization $\bar{\mathcal{E}}^{\text{target}}_{0}=\tau_{\text{global}}/\gamma$.
We then define a dynamic stopping threshold
$\tau_c = \gamma \cdot \bar{\mathcal{E}}^{\text{target}}_{c}$ where $\gamma{=}0.8$.
We stop speculation when the draft entropy $\mathcal{E}_\text{entropy}^{r} < \tau_c$. The threshold thus tightens in complex, high-entropy regions and relaxes in smooth ones, with no extra cost because the target already computes entropy during verification.

\section{Experiments}
\label{sec:experiments}

\subsection{Implementation Details}
\label{sec:impl}
To validate the effectiveness of Drift-AR, we conduct extensive experiments on representative AR-Diffusion hybrids. We cover a broad range of architectures: MAR~\cite{li2024mar}, TransDiff~\cite{zhen2025marrying}, and NextStep-1~\cite{team2025nextstep}. Our method achieves \textbf{1-NFE} for the visual decoding stage: the drift-based decoder performs a single forward pass at inference, without timestep or iterative denoising.

\noindent\textbf{Baselines.}
We compare against the original implementations and state-of-the-art acceleration methods. For AR acceleration: vanilla speculative decoding (SD) and LazyMAR~\cite{yan2025lazymar} (MAR-specific). For diffusion acceleration: standalone DMD~\cite{yin2024one}. All DMD and LazyMAR results use their official configurations; SD uses 6-layer draft models initialized from the target, consistent across baselines.

\noindent\textbf{Drift-AR training.}
We train the drifting field and AR components end-to-end in a single pipeline (no separate CD/DMD stages). For TransDiff: the draft AR has 6 transformer blocks; the drifting decoder replaces the original DiT with our drift-based generator. Phase~I spans 80\% of total epochs ($T_{\text{freeze}}=0.8 \cdot T_{\text{total}}$) with $\alpha$ linearly decaying from $0.95$ to $0$; Phase~II freezes the AR and optimizes only $\mathcal{L}_{\text{drift}}$. Key hyperparameters: $\sigma_{\max}=0.5$, $\tau_\sigma=2.0$, $\mathcal{E}_{\max}$ estimated from training entropy; kernel temperatures $\tau \in \{0.02, 0.05, 0.2\}$. Optimizer: AdamW ($\text{lr}=1\times 10^{-4}$, batch size $32$ for ImageNet). All models are evaluated following the original papers. ImageNet: FID~\cite{heusel2017gans}, IS~\cite{salimans2016improved}, latency on 50,000 images. Text-to-image: GenEval~\cite{ghosh2023geneval}, FID and CLIP on MJHQ-30K~\cite{li2024playground}.

\begin{table}[tb]
\centering
\caption{Class-conditional generation on ImageNet 256$\times$256. Drift-AR achieves the highest speedup while matching or improving FID and IS across all model scales.}
\label{tab:imagenet_gen}
\resizebox{0.6 \linewidth}{!}{%
\begin{tabular}{lllll}
\toprule
\textbf{Method} & \textbf{Latency/s} & \textbf{Speedup} & \textbf{FID $\downarrow$} & \textbf{IS $\uparrow$}\\
\midrule
\multicolumn{5}{l}{\textbf{\textit{MAR}}~\cite{li2024mar}} \\
MAR-L~\cite{li2024mar} & 5.31 & 1.00 & 1.78 & 296.0 \\
MAR-L~\cite{li2024mar} + DMD~\cite{yin2024one} & 1.99 & 2.67 & 1.81 & 295.5 \\
MAR-L~\cite{li2024mar} + LazyMAR~\cite{yan2025lazymar} & 2.29 & 2.32 & 1.93 & 297.4 \\
MAR-L~\cite{li2024mar} + Ours & 0.96 & 5.53 & \textbf{1.76} & \textbf{297.4} \\
MAR-H~\cite{li2024mar} & 9.97 & 1.00 & 1.55 & 303.7 \\
MAR-H~\cite{li2024mar} + DMD~\cite{yin2024one} & 4.17 & 2.39 & 1.73 & 301.0 \\
MAR-H~\cite{li2024mar} + LazyMAR~\cite{yan2025lazymar} & 4.24 & 2.35 & 1.69 & 299.2 \\
MAR-H~\cite{li2024mar} + Ours & 1.93 & 5.16 & \textbf{1.53} & \textbf{304.6} \\
\midrule
\multicolumn{5}{l}{\textbf{\textit{TransDiff}}~\cite{zhen2025marrying}} \\
TransDiff-L~\cite{zhen2025marrying} & 3.17 & 1.00 & 1.61 & 295.1 \\
TransDiff-L~\cite{zhen2025marrying} + SD~\cite{li2024eagle} & 1.75 & 1.81 & 1.88 & 283.6 \\
TransDiff-L~\cite{zhen2025marrying} + DMD~\cite{yin2024one} & 1.61 & 1.97 & 1.79 & 288.3 \\
TransDiff-L~\cite{zhen2025marrying} + Ours & 0.64 & 4.96 & \textbf{1.61} & \textbf{295.8} \\
TransDiff-H~\cite{zhen2025marrying} & 6.72 & 1.00 & 1.55 & 297.9 \\
TransDiff-H~\cite{zhen2025marrying} + SD~\cite{li2024eagle} & 3.52 & 1.91 & 1.68 & 291.1 \\
TransDiff-H~\cite{zhen2025marrying} + DMD~\cite{yin2024one} & 3.48 & 1.93 & 1.71 & 289.9 \\
TransDiff-H~\cite{zhen2025marrying} + Ours & 1.33 & 5.06 & \textbf{1.57} & \textbf{298.1} \\
\bottomrule
\end{tabular}%
}
\end{table}

\subsection{Comparison}
\label{sec:comparison}
\noindent\textbf{ImageNet 256$\times$256 (Class-Conditional).}
As shown in Table~\ref{tab:imagenet_gen}, Drift-AR consistently delivers the highest speedup while maintaining or improving generation quality. For \textbf{MAR}, our method attains 5.53$\times$ and 5.16$\times$ speedup on MAR-L and MAR-H respectively. On MAR-L, we improve both FID (1.76 vs.\ baseline 1.78) and IS (297.4 vs.\ 296.0); on MAR-H, we improve FID (1.53 vs.\ 1.55) and IS (304.6 vs.\ 303.7), with both metrics surpassing DMD and LazyMAR. DMD and LazyMAR achieve only $\sim$2.4$\times$ speedup with degraded or marginal quality. For \textbf{TransDiff}, Drift-AR reaches 4.96$\times$ and 5.06$\times$ speedup on the L and H variants. We improve IS (295.8, 298.1) relative to the baseline (295.1, 297.9). Vanilla SD suffers severe quality collapse (IS drops to 283.6 and 291.1); DMD improves speed but degrades FID/IS. The key advantage is \textbf{1-NFE} visual decoding: our drift-based generator dispenses with iterative denoising entirely, whereas DMD still requires multi-step or distilled sampling.

\begin{figure}[!tbp]
\centering
\includegraphics[width=\linewidth]{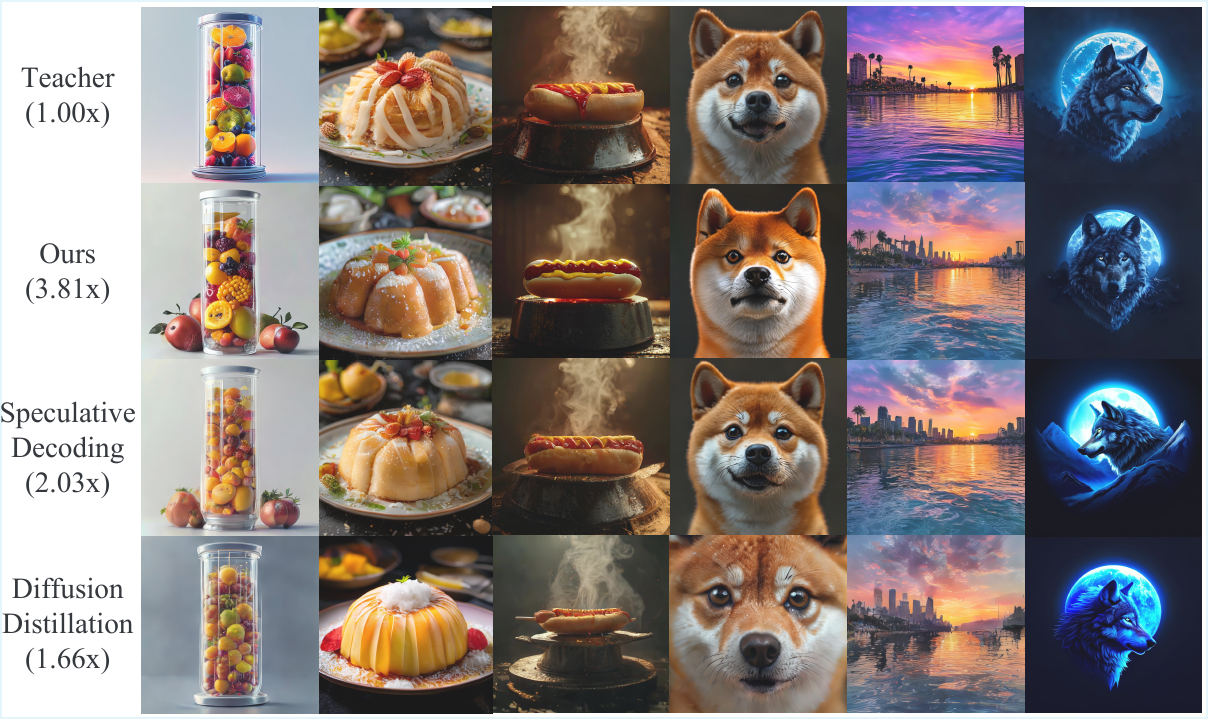}
\caption{Visual comparisons with NextStep-1~\cite{team2025nextstep} on MJHQ-30K~\cite{li2024playground}.}
\label{fig4:vis}
\end{figure}

\noindent\textbf{Text-to-Image (NextStep-1).}
On GenEval and MJHQ-30K (Table~\ref{tab:proprietary_gen}), Drift-AR achieves 3.81$\times$ speedup with GenEval 0.66, FID 6.66, and CLIP 29.02---all superior to the baseline (0.63, 6.71, 28.67). SD reaches 2.03$\times$ but loses quality (FID 6.90, CLIP 27.96); DMD collapses (FID 8.33, CLIP 25.19). Qualitative comparisons in Fig.~\ref{fig4:vis} confirm that our method preserves fine-grained details and semantic consistency while dramatically reducing latency.

\subsection{Ablation Study}
We ablate five key components on TransDiff-H, with all other settings consistent with our full framework. Refer to the supplementary material for more details.

\noindent\textbf{$\mathbb{A}$: Ablating Entropy Parameterization.} We replace the entropy-parameterized variance $\sigma^2(\mathcal{E}_{entropy})$ with a constant variance for the initial drifting state, testing whether entropy as the physical variance is essential for the drifting field's equilibrium.

\noindent\textbf{$\mathbb{B}$: Ablating Stage-Wise Loss Reweighting.} We fix $\alpha = 0.5$ throughout training instead of the dynamic annealing $\alpha(t)$, examining whether adaptive reweighting is necessary to coordinate AR and drifting-field components.

\noindent\textbf{$\mathbb{C}$: Ablating the Anti-Symmetric Kernel Objective.} We keep only the \emph{attraction} term and remove the \emph{repulsion} term (non-antisymmetric variant). The resulting field no longer satisfies $q{=}p \Rightarrow V{=}0$, so drift need not vanish at equilibrium.

\noindent\textbf{$\mathbb{D}$: Ablating Entropy-Based Early Stopping in Inference.} We disable entropy monitoring during speculative decoding, assessing whether the early stopping mechanism mitigates residual low-quality draft output.

\noindent\textbf{$\mathbb{E}$: Ablating Prior Freezing in Phase~II.} We allow $\mathcal{L}_{\text{drift}}$ gradients to back-propagate through the AR module in Phase~II (i.e., we do not detach $z_{AR}$ and $\mathbf{E}$), testing the ``moving prior'' hypothesis.

\begin{table}[t]
    \caption{(left) Text-to-image comparison on GenEval and MJHQ-30K. (right) Ablation study on TransDiff-H (ImageNet 256$\times$256): removing entropy parameterization ($\mathbb{A}$) causes the largest degradation, validating entropy as the core design.}
    \label{tab:proprietary_gen}
    \begin{subtable}{.6\linewidth}
      \centering
        \resizebox{!}{0.9cm}{
        \begin{tabular}{lcccc}
            \toprule
            \textbf{Method} & \textbf{Speedup} & \textbf{GenEval $\uparrow$} & \textbf{FID $\downarrow$} & \textbf{CLIP $\uparrow$}\\
\midrule
NextStep-1~\cite{team2025nextstep} & 1.00 & 0.63 & 6.71 & 28.67 \\
NextStep-1~\cite{team2025nextstep} + SD~\cite{li2024eagle} & 2.03 & 0.63 & 6.90 & 27.96 \\
NextStep-1~\cite{team2025nextstep} + DMD~\cite{yin2024one} & 1.66 & 0.59 & 8.33 & 25.19 \\
NextStep-1~\cite{team2025nextstep} + Ours & 3.81 & \textbf{0.66} & \textbf{6.66} & \textbf{29.02} \\
\bottomrule
        \end{tabular}
        }
    \end{subtable}%
    \begin{subtable}{.4\linewidth}
      \centering
        \resizebox{!}{1.2cm}{
        \begin{tabular}{lcc}
            \toprule
\textbf{Method} & \textbf{FID $\downarrow$} & \textbf{IS $\uparrow$}\\
\midrule
TransDiff-H & 1.55 & 297.9 \\
Ours w/o $\mathbb{A}$ & 1.72 & 289.3 \\
Ours w/o $\mathbb{B}$ & 1.69 & 292.6 \\
Ours w/o $\mathbb{C}$ & 1.69 & 291.8 \\
Ours w/o $\mathbb{D}$ & 1.62 & 295.5 \\
Ours w/o $\mathbb{E}$ & 1.67 & 291.9 \\
\midrule
Ours (full) & \textbf{1.57} & \textbf{298.1} \\
\bottomrule
        \end{tabular}
        }
    \end{subtable} 
\end{table}

\noindent\textbf{Analysis of Ablation Results (Table~\ref{tab:proprietary_gen}).}
Removing any component degrades both FID and IS. \textit{w/o $\mathbb{A}$} (entropy parameterization) produces the largest degradation: FID rises to 1.72 and IS drops to 289.3. This is the most critical ablation, as it directly validates the core thesis that entropy should parameterize the drifting prior: without entropy-driven variance, the drifting field receives a fixed-variance prior that cannot adapt to local generation difficulty (cf.\ Fig.~\ref{fig2:moti}(c)(d)). \textit{w/o $\mathbb{B}$} (stage-wise reweighting): FID 1.69, IS 292.6. Fixing $\alpha{=}0.5$ prevents the AR prior from stabilizing before Phase~II; the drifting field is trained against a moving prior, undermining the equilibrium condition. \textit{w/o $\mathbb{C}$} (anti-symmetric kernel): FID 1.69, IS 291.8. The non-antisymmetric variant lacks the guarantee $q{=}p \Rightarrow V{=}0$, leading to residual drift at convergence and lower-quality 1-NFE samples. \textit{w/o $\mathbb{D}$} (early stopping): FID 1.62, IS 295.5. Disabling context-aware early stopping allows low-quality or overconfident draft outputs to propagate, increasing rejection overhead and slightly degrading diversity. \textit{w/o $\mathbb{E}$} (prior freezing): FID 1.67, IS 291.9. Allowing drift gradients to update the AR in Phase~II creates a moving prior that destabilizes the drifting field; the equilibrium condition $q{=}p \Rightarrow V{=}0$ is never satisfied stably, leading to noticeable quality degradation. The full model (1.57 FID, 298.1 IS) confirms that all five components contribute to the final performance.

\subsection{Analysis}
\label{sec:analysis}

\noindent\textbf{Decoder Step Analysis.}
A natural question is whether the drifting decoder truly needs only one forward pass. Table~\ref{tab:nfe} compares three decoder types on TransDiff-H at varying step counts. TransDiff uses 20 denoising steps by default. The original decoder degrades rapidly when fewer steps are used, collapsing to FID 14.72 at 1 step. DMD recovers quality through distillation but still yields FID 2.93 at 1 step. In contrast, our drifting decoder achieves FID 1.57 at 1 step---on par with the 20-step diffusion baseline---confirming that the anti-symmetric drifting field concentrates the generative capacity into a single pass.

\begin{table}[tb]
\centering
\caption{Decoder step analysis on TransDiff-H (ImageNet 256$\times$256). Our drifting decoder at 1 step matches the 20-step diffusion baseline, while DMD still degrades significantly at 1 step.}
\label{tab:nfe}
\resizebox{0.6 \linewidth}{!}{%
\begin{tabular}{llrrrr}
\toprule
\textbf{Decoder} & \textbf{Steps} & \textbf{Latency/s} & \textbf{Speedup} & \textbf{FID $\downarrow$} & \textbf{IS $\uparrow$}\\
\midrule
Diffusion & 20 (default) & 6.72 & 1.00$\times$ & 1.55 & 297.9 \\
Diffusion & 8 & 2.85 & 2.36$\times$ & 2.34 & 279.8 \\
Diffusion & 4 & 1.65 & 4.07$\times$ & 3.89 & 261.5 \\
Diffusion & 1 & 1.28 & 5.25$\times$ & 14.72 & 148.3 \\
\midrule
DMD~\cite{yin2024one} & 4 & 1.85 & 3.63$\times$ & 2.11 & 282.9 \\
DMD~\cite{yin2024one} & 1 & 1.30 & 5.17$\times$ & 2.93 & 273.2 \\
\midrule
Drifting (Ours) & 1 & 1.33 & 5.06$\times$ & \textbf{1.57} & \textbf{298.1} \\
\bottomrule
\end{tabular}%
}
\end{table}

\noindent\textbf{Sensitivity to $\sigma_{\max}$.}
The maximum variance $\sigma_{\max}$ in Eq.~\ref{eq:sigma_mapping} controls the range of entropy-to-variance mapping. Table~\ref{tab:sigma} shows that performance is robust across a moderate range ($0.3$--$0.7$), peaking at $\sigma_{\max}{=}0.5$. Too small a value ($0.1$) collapses the prior toward a Dirac delta, depriving the drifting field of sufficient randomness to cover the data manifold. Too large a value ($1.0$) produces an overly diffuse prior that demands excessive drift, degrading quality.

\begin{table}[tb]
\centering
\caption{Sensitivity to $\sigma_{\max}$ on TransDiff-H (ImageNet 256$\times$256). Performance is robust across $0.3$--$0.7$; extremes hurt by under- or over-spreading the entropy-parameterized prior.}
\label{tab:sigma}
\begin{tabular}{cccccc}
\toprule
$\sigma_{\max}$ & 0.1 & 0.3 & \textbf{0.5} & 0.7 & 1.0\\
\midrule
\textbf{FID $\downarrow$} & 1.68 & 1.58 & \textbf{1.57} & 1.57 & 1.64\\
\textbf{IS $\uparrow$} & 292.7 & 296.5 & \textbf{298.1} & 296.3 & 293.1\\
\bottomrule
\end{tabular}
\end{table}

\noindent\textbf{Entropy--Prediction Error Correlation.}
Fig.~\ref{fig2:moti}(c)(d) quantifies the relationship between per-position AR entropy $\mathcal{E}^{(r)}$ and prediction error $\|z_{AR}^{(r)} - x_{gt}^{(r)}\|$ on the ImageNet validation set. The scatter plot shows a clear positive correlation (Pearson $r{=}0.64$), and the binned analysis confirms that mean error increases monotonically with entropy. This validates our entropy-parameterized variance: positions where the AR model is uncertain are precisely those where the decoder must apply the strongest correction, and our mapping (Eq.~\ref{eq:sigma_mapping}) allocates prior spread accordingly.

\FloatBarrier
\section{Discussion and Conclusion}
\label{sec:discussion}
We show that the per-position prediction entropy of continuous-space AR models, a signal previously overlooked, simultaneously governs draft quality in speculative decoding and encodes local generation difficulty for visual decoding. This dual role makes entropy the natural unifying principle for accelerating AR-Diffusion hybrids. Drift-AR exploits this by (1)~aligning draft--target entropy distributions for efficient speculation, and (2)~reinterpreting entropy as the physical variance of an anti-symmetric drifting field, achieving 1-NFE visual decoding without distillation. Experiments on MAR, TransDiff, and NextStep-1 demonstrate 3.8--5.5$\times$ speedup while matching or surpassing original quality.

\noindent\textbf{Limitations.} The entropy estimation relies on attention-map statistics, whose quality may degrade under distribution shift or for architectures without causal attention. The two-phase annealed training requires a longer schedule than standard distillation to stabilize the prior before drifting-field optimization. Extending the framework to video or higher-resolution generation, where spatial entropy patterns become more complex, remains future work.

\bibliographystyle{splncs04}
\bibliography{main}
\end{document}